\documentclass[conference]{IEEEtran}
\IEEEoverridecommandlockouts

\usepackage{cite}
\usepackage{amsmath,amssymb,amsfonts}
\usepackage{algorithmic}
\usepackage{graphicx}
\usepackage{textcomp}
\usepackage{xcolor}
\usepackage{url} 
\usepackage[T1]{fontenc}
\usepackage[utf8x]{inputenc}
\usepackage{booktabs}
\usepackage{tabularx}
\usepackage{comment}

\def\BibTeX{{\rm B\kern-.05em{\sc i\kern-.025em b}\kern-.08em
    T\kern-.1667em\lower.7ex\hbox{E}\kern-.125emX}}
\begin{document}

\title{A Small-Scale Robot for Autonomous Driving: Design, Challenges, and Best Practices\\
\thanks{This work was partially supported by the National Science Foundation
under Grant CNS-1932037. The code and supplementary material are available
at: https://github.com/Hmaghsoumi/F1SIXTH.}
}

\author{\IEEEauthorblockN{Hossein Maghsoumi, Yaser Fallah}
\IEEEauthorblockA{\textit{Department of Electrical and Computer Engineering} \\
\textit{University of Central Florida}\\
Orlando, USA \\
hossein.maghsoumi@ucf.edu, yaser.fallah@ucf.edu}
}

\maketitle

\IEEEpubid{%
  \begin{minipage}{\textwidth}
    \vspace{6.8\baselineskip}   
    \centering
    \fbox{%
      \parbox{0.92\textwidth}{\centering\small
        This work has been submitted to the IEEE for possible publication.
        Copyright may be transferred without notice, after which this
        version may no longer be accessible.}%
    }%
  \end{minipage}%
}%
\IEEEpubidadjcol

\begin{abstract}
Small-scale autonomous vehicle platforms provide a cost-effective environment for developing and testing advanced driving systems. However, specific configurations within this scale are underrepresented, limiting full awareness of their potential. This paper focuses on a one-sixth-scale setup, offering a high-level overview of its design, hardware and software integration, and typical challenges encountered during development. We discuss methods for addressing mechanical and electronic issues common to this scale and propose guidelines for improving reliability and performance. By sharing these insights, we aim to expand the utility of small-scale vehicles for testing autonomous driving algorithms and to encourage further research in this domain.

\end{abstract}


\section{Introduction}
Autonomous driving is the ability of a vehicle to perceive its surroundings, interpret that information, and execute safe control actions without direct human intervention \cite{10137425}.
Autonomous driving research has benefited significantly from scaled-down platforms, which offer a safer and more cost-effective environment for rapid prototyping and experimentation \cite{burns2024openconvoy}. A popular example is the F1TENTH project, where a one-tenth-scale car is equipped with essential hardware and software to explore various autonomous navigation techniques. However, despite the success of such a platform, there are other small-scale options—such as the F1SIXTH platform—that are larger than F1TENTH yet still rely on relatively inexpensive materials and can be deployed in a wide range of autonomous vehicle projects. By operating at a slightly larger scale, F1SIXTH more closely approximates the dynamics of a full-size car while preserving many advantages of small-scale research, including reduced risk and lower operating costs, Fig. 1.

While the F1TENTH platform has garnered extensive research attention and documentation, other small-scale platforms—such as F1SIXTH—remain relatively underrepresented. This lack of published studies means that the valuable insights and distinct advantages offered by these other small-scale vehicles often go unnoticed or underutilized. In particular, their increased size can better approximate full-scale car dynamics without sacrificing the core benefits of small-scale experimentation. However, with minimal research and insufficient documentation, the potential of platforms like F1SIXTH risks being overshadowed by the established prominence of F1TENTH, leaving an important gap in autonomous vehicle experimentation resources.

\begin{figure}[t]
\centerline{\includegraphics[width=1.0\linewidth]{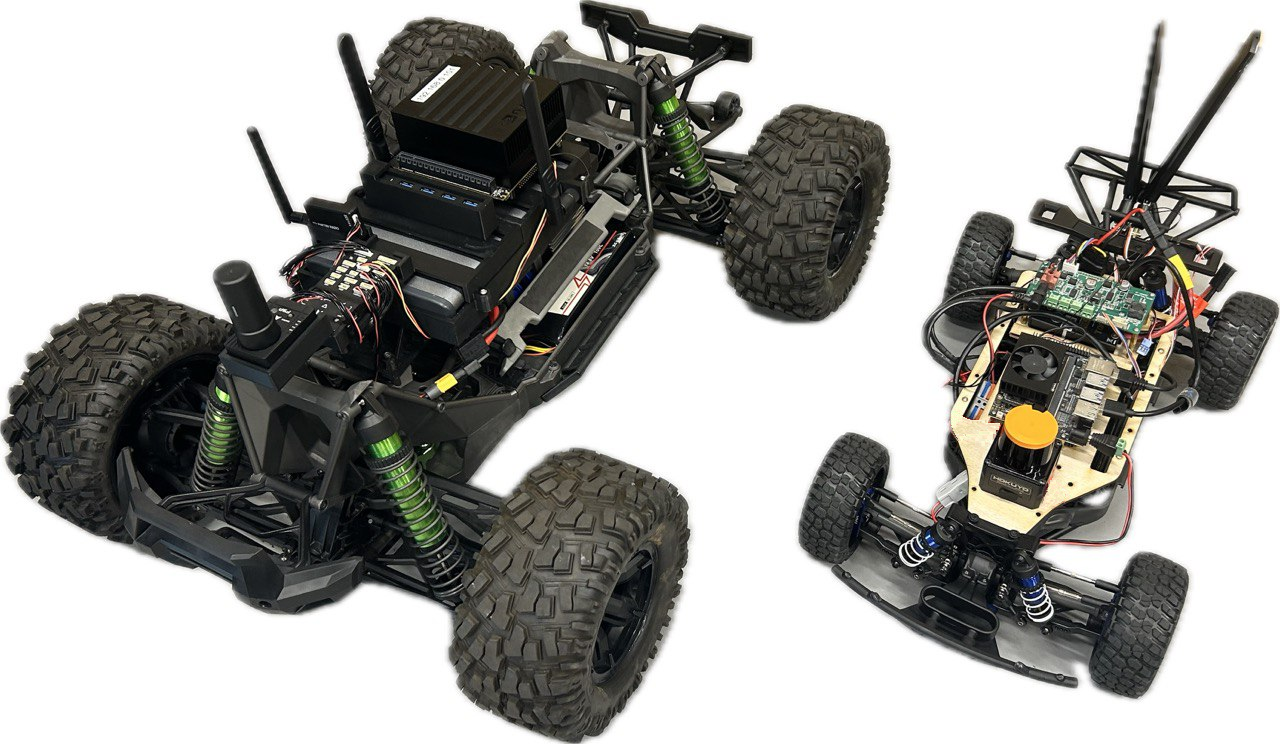}}
\caption{Visual Contrast Between F1SIXTH and F1TENTH Testbeds}
\label{fig:Compare}
\end{figure}

In this paper, we seek to address this gap by providing a detailed overview of F1SIXTH’s design, the challenges encountered, and the corresponding solutions implemented. We outline the vehicle’s mechanical configuration and discuss how it accommodates a slightly larger form factor while retaining the practical advantages of smaller platforms. We also touch upon the electronic integration and software stack that enable autonomous functionality \cite{Mygithub}. By sharing our experiences, we hope to illuminate the untapped potential of F1SIXTH for autonomous vehicle research and inspire others to explore and advance this platform beyond the well-established F1TENTH ecosystem.

The remainder of this paper is organized as follows. First, we present a brief background on relevant scaled autonomous vehicle platforms and highlight the unique features of F1SIXTH. We then detail the major hardware and software components of our system, followed by a discussion of the key challenges we faced and the corresponding solutions. Finally, we provide a concise step-by-step setup guide and conclude with reflections on the efficacy of F1SIXTH for autonomous research, along with suggestions for future work.

\section{Background and Related Work}
Scaled autonomous vehicle platforms play a pivotal role in academic and industrial research, offering a controlled testbed for evaluating algorithms related to perception, planning, and control. By reducing vehicle dimensions and power, researchers can rapidly prototype and iterate on their designs without incurring the high costs and risks of full-scale experiments. Among these platforms, F1TENTH has emerged as a widely recognized standard, supported by a substantial community of practitioners who share open-source hardware and software resources \cite{f1tenth}, \cite{9216949}. F1TENTH’s one-tenth-scale race car design enables an approachable entry point for students, researchers, and enthusiasts to experiment with autonomy in a structured, documented environment \cite{baumann2024forzaeth}.

Nevertheless, the spectrum of scaled vehicles extends well beyond F1TENTH. The Audi Autonomous Driving Cup (AADC) \cite{audiweb}, for instance, employs 1:8-scale model cars in an international student competition, requiring participants to develop robust perception and planning solutions in a standardized, challenge-based format \cite{kuhnt2016robust}. Duckietown \cite{paull2017duckietown} offers an equally engaging but more compact platform, using off-the-shelf components and a monocular camera for onboard sensing. Its focus on affordability and open-source collaboration has proven valuable for educational settings. Meanwhile, JetRacer \cite{jetracerweb}—an AI-oriented race car developed by NVIDIA—comes in both 1:18 and 1:10 variations, pairing rapid prototyping with real-time computer vision and machine learning capabilities \cite{reviewresearchav}. Community-driven projects like Donkeycar \cite{donkeycar}, NXPCup \cite{nxpcupweb}, and ROAR \cite{roarweb2} similarly highlight the versatility of small-scale setups by providing customizable hardware and software stacks aimed at tasks ranging from lane-following to autonomous racing. Although each of these platforms addresses distinct educational and research goals, no single scale or configuration dominates every experimental need. Larger small-scale models can capture vehicle dynamics that more closely resemble those of real cars, offering superior stability and road-grip characteristics, all while maintaining a manageable cost and risk for laboratory use.

This is precisely where F1SIXTH finds its niche: it occupies a slightly bigger footprint than F1TENTH—making it better suited for heavier payloads, longer battery life, and more realistic driving dynamics—yet it retains much of the operational accessibility that smaller robotic platforms provide. Despite these advantages, relatively few academic publications or comprehensive build guides have focused on F1SIXTH, leaving a notable gap in available documentation. By consolidating information about F1SIXTH’s hardware components, software stack, and operating procedures, this work aims to foster broader adoption and spark further research on this promising intermediate-scale platform.

\section{System Design}
The F1SIXTH is typically built around a Traxxas X-Maxx chassis, which uses a high-strength molded composite for rigidity and impact resistance. Equipped with a single powerful brushless motor, multiple batteries, and a suite of sensors, this platform can be configured for diverse autonomous driving experiments. An example of our assembled F1SIXTH setup is shown in Fig. 2.

\subsection{Mechanical Structure} 
The chassis is designed to handle moderate- to high-speed operations and absorb shocks from uneven surfaces or minor collisions. Key mechanical elements include:
\begin{itemize}
    \item Chassis Frame: Often constructed from high-strength composite or aluminum, providing space for mounting additional hardware.
    \item Motor: Drives the drivetrain; typically a brushless model capable of delivering high torque and top speed suitable for autonomous driving experiments.
    \item Steering Mechanism: A servo-controlled front axle converts control signals into steering angles.
    \item Suspension System and Wheels: Calibrated for stability in higher-speed maneuvers. The suspension setup includes adjustable springs, allowing users to fine-tune damping and stiffness depending on their testing environment.
    
\end{itemize}

\subsection{Electronic Components}
To enable autonomous or semi-autonomous control, F1SIXTH relies on a range of electronic modules:

\begin{itemize}
    \item Autopilot Unit (e.g., Pixhawk): Executes control loops (steering, throttle) and handles sensor fusion.
    \item Onboard Processor (e.g., NVIDIA Jetson): Manages higher-level tasks (e.g., vision, mapping, path planning).
    \item Electronic Speed Controller (ESC): Supplies power to the motor at varying levels, facilitating speed control.
    \item Battery System: Powers all onboard electronics, typically consisting of multiple LiPo packs for extended runtime, along with a separate battery dedicated to the onboard processor. 
    \item Telemetry \& Network Modules: Provide remote data access (e.g., SIK radio, Wi-Fi, 4G/5G adapters)
\end{itemize}

\begin{figure*}[t]
\centerline{\includegraphics[width=1.0\linewidth]{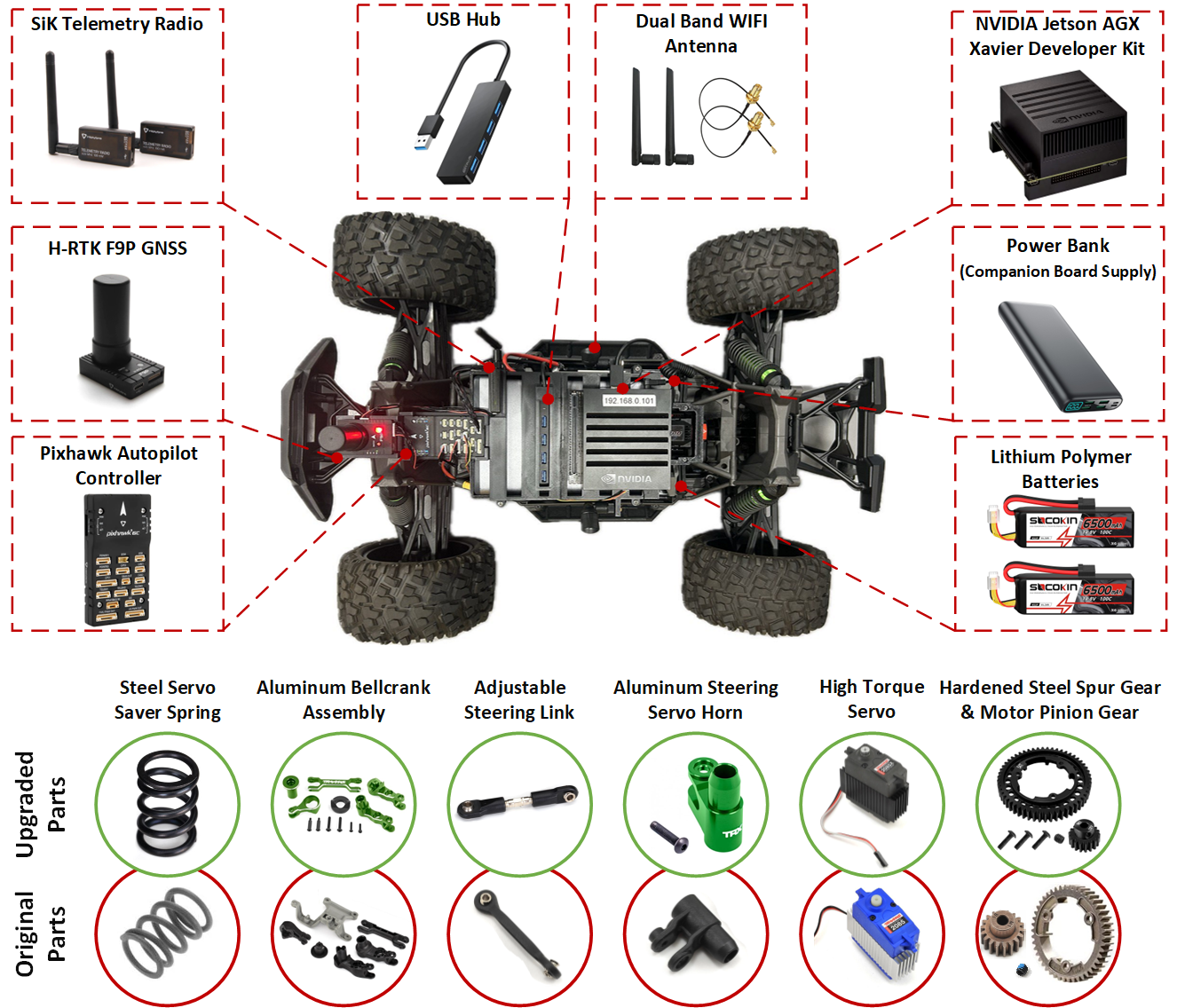}}
\caption{Overview of the F1sixth Vehicle Setup}
\label{Vehicle_Setup}
\end{figure*}

\subsection{Sensor Suite}

The platform can be equipped with a range of sensors based on project requirements:

\begin{itemize}
    \item Wheel Encoders: Improve odometry accuracy by measuring wheel rotation.
    \item IMU: Tracks orientation and acceleration, essential for state estimation.
    \item GPS: Useful for outdoor localization.
    \item LiDAR \& Cameras: Useful for advanced perception tasks \cite{7395239}, \cite{7125330} such as obstacle detection \cite{10944753}, lane following \cite{10385062}, or SLAM.
\end{itemize}

\subsection{Software Stack}
The software used on F1SIXTH combines both low-level firmware and high-level autonomy:
\begin{itemize}
\item Autopilot Firmware (e.g., ArduPilot, PX4): Manages sensor fusion (IMU, GPS, etc.) and executes control loops for stable navigation.

\item Robot Operating System (ROS): Coordinates communication between various software nodes for perception, path planning, and control.

\item Custom Control \& Perception Algorithms: Depending on project goals, additional modules or machine learning models can be integrated to handle tasks like obstacle avoidance or lane detection.
\end{itemize}

\subsection{Integration Considerations}

A few best practices help ensure reliable system performance:
\begin{itemize}
\item Wiring \& Connectors: Use labeled, secure connectors to avoid loose cables and signal dropouts.

\item Power Distribution: Isolate sensitive components with separate voltage rails or regulators if possible.

\item Thermal Management: Provide adequate airflow or heatsinking for components like the Jetson.
\end{itemize}

\subsection{Common Challenges and Suggested Solutions}
Despite its robust design, F1SIXTH can encounter specific issues during high-speed operations or repeated testing. Below are some common challenges and practical methods to address them:

\subsubsection{Servo Saver Spring Fatigue}

\begin{itemize}
    \item \textbf{Challenge:} Repeated stress from collisions weakens the servo saver spring, reducing steering stability.
    \item \textbf{Solution:} Periodically inspect this spring; upgrade to higher-quality steel springs if you notice any loss in steering stiffness.
\end{itemize}

\subsubsection{Front Wheel Calibration Limitations}

\begin{itemize}
    \item \textbf{Challenge:} The original steering components may not allow fine-tuned wheel alignment, causing steering drift or uneven tire wear.
    \item \textbf{Solution:} Install an adjustable steering link (separately sold by the manufacturer) to properly align the front wheels and improve handling.
\end{itemize}

\subsubsection{Gear \& Axle Carrier Breakage}

\begin{itemize}
    \item \textbf{Challenge:} Collisions or abrupt impacts can crack or strip the gear (e.g., pinion/spur) or axle carriers.
    \item \textbf{Solution:} Stock spare gears and axle carriers. Consider upgrading to metal gears if testing involves frequent high-impact maneuvers.
\end{itemize}

\subsubsection{Lubrication and Maintenance}

\begin{itemize}
    \item \textbf{Challenge:} Heavier vehicles subject gears, bearings, and shafts to increased stress, especially during extended or off-road testing.
    \item \textbf{Solution:} Apply specialized grease or oils to driveline components. Routine lubrication extends part life and ensures smoother performance.
\end{itemize}

\subsubsection{Servo and Steering Upgrades}

\begin{itemize}
    \item \textbf{Challenge:} Stock servo and plastic steering assemblies can limit steering precision and fail under heavy loads.
    \item \textbf{Solution:} Upgrade to a high-torque, waterproof digital servo and use an aluminum bellcrank or servo horn for improved durability and handling responsiveness.
\end{itemize}

\section{Customizations and Configurations}
This section highlights several enhancements to the baseline F1SIXTH platform: a custom holder for mounting key electronics, firmware settings in Autopilot, and an overview of upgraded parts and their associated costs.

\subsection{Holder Design}
Although the Traxxas X-Maxx chassis provides ample space for off-the-shelf equipment, certain components—such as the companion computer (Jetson), autopilot (Pixhawk), and GPS module (F9P)—benefit from a dedicated mounting system that keeps wiring organized, minimizes vibration, and simplifies future alterations. To address these needs, we designed a 3D-printed holder with the following features, Fig. 3.
\begin{itemize}
    \item Processor Compartment: Secures the Jetson at a stable height, providing adequate airflow and straightforward cable routing.
    \item Autopilot Slot: Houses a Pixhawk (or similar flight controller) on stable supports to reduce IMU interference and noise.
    \item Sensor Bays: Allocates space for the F9P GPS module and any telemetry radios, ensuring minimal electromagnetic interference from the drivetrain.
    \item Optional Extension Module: Allows the autopilot or GPS to be mounted near the back axis. Some control algorithms (e.g., Pure Pursuit) operate more accurately when the GPS is placed closer to the rear axle, while others (e.g., Stanley) may prefer it near the front. This modular approach lets researchers easily switch configurations without extensive modifications to the chassis.
\end{itemize}

An STL file for the holder can be found in our GitHub repository at \cite{Mygithub}. In practice, the holder’s alignment and mounting holes are designed to integrate seamlessly onto the X-Maxx frame with only minor drilling or by repurposing existing bolt locations.

\begin{figure}[t]
\centerline{\includegraphics[width=1.0\linewidth]{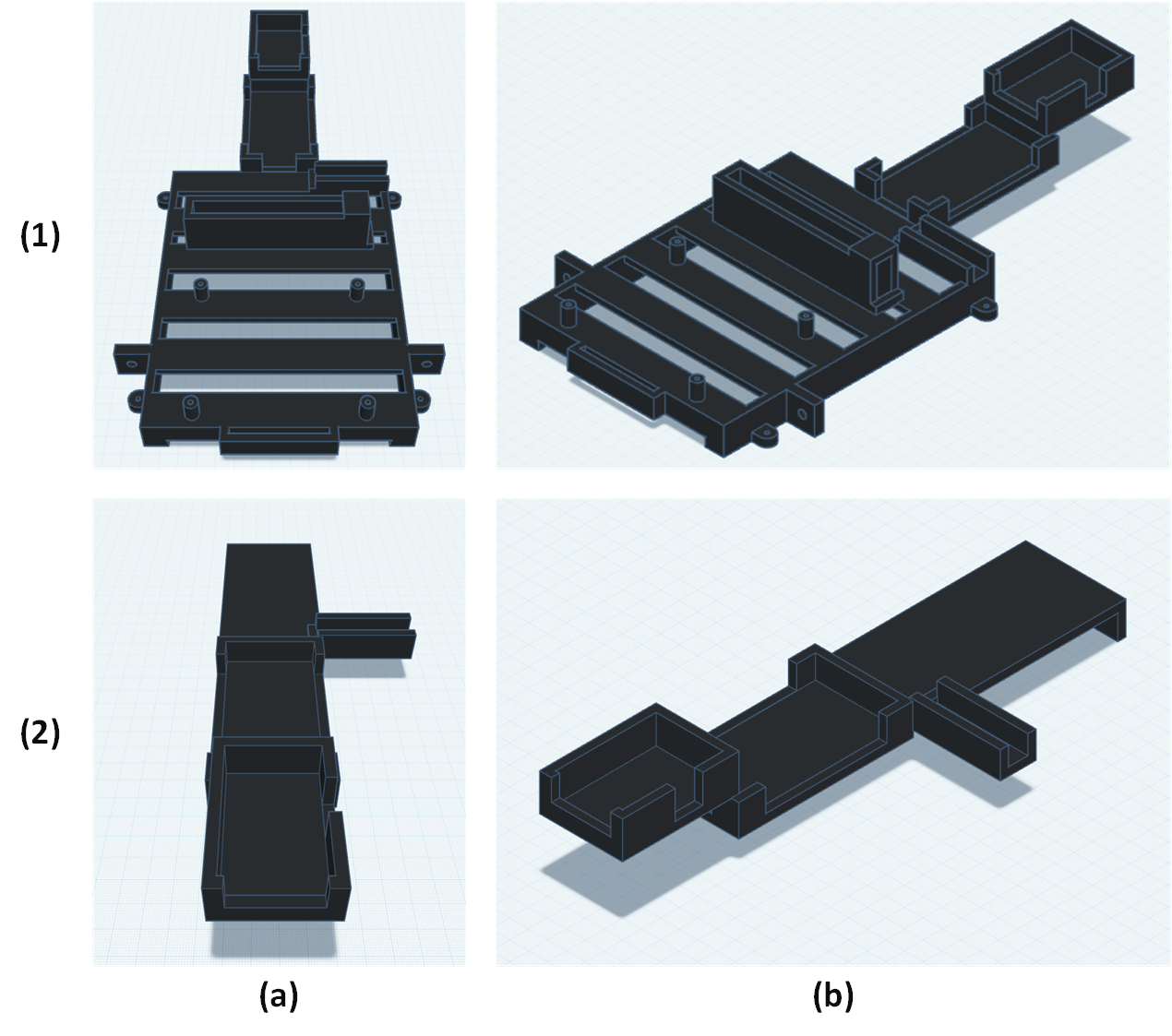}}
\caption{3D-Printed Holder Design for Onboard Electronics: (1) Main holder and (2) optional extension module, each shown in (a) perspective view and (b) flat view.}
\label{3DPrinted}
\end{figure}

\subsection{Firmware Parameter Configuration}
\label{sec:firmware-params}

PX4 and ArduPilot are widely used autopilot firmwares that support ground rover functionality. 
Several key PX4 Ground Rover parameters were tailored to the physical traits of F1SIXTH, including speed limits, throttle range, GPS communication settings, and PWM outputs for steering and throttle. Table~\ref{tab:px4-params-table} summarizes these parameters, their ArduPilot equivalents, and concise notes on their purpose. Through practical testing, we observed that fine-tuning throttle limits, wheelbase dimensions, and GPS baud rates significantly impacts navigational stability and responsiveness. Ensuring consistency between parameter settings and real-world measurements (e.g., wheelbase) was critical for accurate trajectory following.

\begin{table*}[ht]
\centering
\caption{PX4 Parameter Summary for the F1SIXTH Platform (with ArduPilot Equivalents)}
\label{tab:px4-params-table}
\begin{tabularx}{\textwidth}{l c l X}
\toprule
\textbf{PX4 Parameter} & \textbf{Value} & \textbf{ArduPilot Equivalent} & \textbf{Description (expanded)} \\
\midrule
\multicolumn{4}{l}{\textit{Speed and Throttle}}\\
\midrule
GND\_SPEED\_IMAX    & 0.125\,\%/m/s & SPEED\_IMAX   & Integrator windup limit for the ground-speed PID loop.                                  \\
GND\_SPEED\_P       & 0.250\,\%/m/s & SPEED\_P      & Proportional gain that determines how aggressively throttle responds to speed error.     \\
GND\_SPEED\_THR\_SC & 1.000\,\%/m/s & SPEED\_SCALER & Scaling factor that converts controller output to a throttle percentage.                 \\
GND\_THR\_CRUISE    & 30.0\,\%      & THR\_CRUISE   & Nominal throttle used for steady, level driving.                                         \\
GND\_THR\_MAX       & 50.0\,\%      & THR\_MAX      & Hard cap on throttle; prevents aggressive acceleration that could break traction.        \\
GND\_THR\_MIN       & 0.0\,\%       & THR\_MIN      & Minimum throttle; keeps motor from stalling during low-speed maneuvers.                 \\
GND\_WHEEL\_BASE    & 1.575\,ft     & WHEEL\_BASE   & Physical distance between front and rear axles; critical for path-tracking geometry.    \\
\midrule
\multicolumn{4}{l}{\textit{GPS}}\\
\midrule
GPS\_UBX\_BAUD2     & 115200\,B/s   & GPS\_BAUD       & High-speed UART baud rate for u-blox receiver, enabling 5–10 Hz RTK updates.            \\
GPS\_UBX\_DYNMODEL  & automotive    & GPS\_NAVFILTER  & Sets Kalman filter assumptions for a wheeled vehicle (low-altitude, moderate dynamics). \\
GPS\_UBX\_MODE      & Rover+Base    & GPS\_TYPE       & Configures RTK rover with corrections from a static base station for centimeter accuracy.\\
\midrule
\multicolumn{4}{l}{\textit{MAVLink and Serial}}\\
\midrule
MAV\_1\_CONFIG      & TELEM2        & SERIALn\_PROTOCOL=1 & Routes MAVLink stream to the second telemetry port.                                 \\
MAV\_1\_RATE        & 10000 B/s   & SERIALn\_BAUD       & Payload rate matched to 115200-baud link \\

MAV\_TYPE           & Ground Rover  & FRAME\_CLASS=ROVER  & Announces vehicle class so GCS and companion code apply rover-specific logic.          \\
\midrule
\multicolumn{4}{l}{\textit{PWM / Servo Outputs}}\\
\midrule
PWM\_MAIN\_DIS2     & 1500          & SERVOx\_TRIM & Neutral PWM when disarmed (steering channel).                                            \\
PWM\_MAIN\_DIS7     & 1500          & SERVOx\_TRIM & Neutral PWM when disarmed (throttle channel).                                            \\
PWM\_MAIN\_FUNC2    & Steering      & SERVOx\_FUNCTION=26 & Maps MAIN 2 to ground-steering output.                                               \\
PWM\_MAIN\_FUNC7    & Throttle      & SERVOx\_FUNCTION=70 & Maps MAIN 7 to throttle/ESC output.                                                  \\
\midrule
\multicolumn{4}{l}{\textit{Serial Baudrates}}\\
\midrule
SER\_GPS1\_BAUD     & 115200\,8N1 & SERIALn\_BAUD & Primary GPS port—high baud for RTK.                                                     \\
SER\_TEL2\_BAUD     & 115200\,8N1 & SERIALn\_BAUD & Secondary telemetry—matches companion computer link rate.                               \\
\bottomrule
\end{tabularx}
\end{table*}

\begin{table*}[ht]
\centering
\caption{Upgraded Components for F1sixth Platform}
\begin{tabularx}{\textwidth}{lXll}
\toprule
\textbf{Item} & \textbf{Upgrade Description} & \textbf{Approx. Cost (USD)} & \textbf{Rationale} \\
\midrule
Metal Gears (Pinion/Spur) & Hardened steel or aluminum & 20--30 & Resists breakage during collisions \\
Servo Saver Spring & High-quality steel spring & 5--8 & Maintains steering firmness under stress \\
High-Torque Digital Servo & Waterproof, metal gears & 90--100 & Improves steering precision and durability \\
Aluminum Steering Assembly & Metal bellcrank, servo horn & 75--95 & Minimizes steering slop and plastic part failures \\
Adjustable Steering Link & Threaded rods with turnbuckles & 5--8 & Allows precise front-wheel alignment \\
Custom Electronics Holder & 3D-printed design (GitHub STL) & 10--40 (material cost) & Organizes Jetson, Pixhawk, GPS, and other modules \\
\bottomrule
\end{tabularx}
\label{tab:upgrades}
\end{table*}

Raising \texttt{GND\_THR\_MAX} above 50\,\% improves acceleration but risks abrupt torque onset at low speeds. Ensuring \texttt{GND\_WHEEL\_BASE} accurately matches the physical axle separation is also vital for precise turning. Overall, proper tuning of these parameters yields smoother motion and a more reliable control experience for F1SIXTH on both PX4 and ArduPilot.

\subsection{Upgraded Components and Cost Overview}
During assembly and testing, various upgrades—from heavy-duty gears to high-torque servos—proved essential for reliability and performance under demanding conditions. Table~\ref{tab:upgrades} summarizes these enhancements, including approximate costs. Depending on the source or vendor, prices may vary.

Investing in these parts helps the F1SIXTH endure high-speed maneuvers, occasional crashes, and extended test sessions—common stresses in autonomous vehicle research. By consolidating data on recommended upgrades and essential parameter configurations, we aim to streamline the setup process and encourage broader adoption of this intermediate-scale platform.

\section{Experimental Validation}
\subsection{Testbed and Instrumentation}
Experiments were performed on three identical F1SIXTH vehicles equipped with

\begin{itemize}
    \item a Pixhawk-based drive-by-wire stack (steering + throttle)
    \item an NVIDIA Jetson companion computer running ROS 2
    \item F9P RTK-GNSS for 1-2 cm localisation
    \item IEEE 802.11ac radios for inter-vehicle V2V messaging
\end{itemize}

All vehicles used the hardware upgrades and PX4 parameters already detailed in Section IV to ensure mechanical robustness and precise control. The complete ROS launch files, log bags, and CAD of the 3-D-printed holder are available in the project repository \cite{Mygithub}, \cite{Mygithub2}.

\subsection{Closed-Loop Trajectory-Tracking Accuracy}
A single-vehicle baseline run established control fidelity.  
The leader drove an \(8\,\text{m}\times 4\,\text{m}\) oval at two speeds (1 m s\(^{-1}\) for the first half-lap, 2 m s\(^{-1}\) thereafter).  
The cross-track root-mean-square error (RMSE) never exceeded \(\mathbf{0.07}\,\text{m}\), confirming that the Stanley lateral controller and PID speed loop are correctly tuned for the heavier 1/6 chassis. 

\begin{figure}[t]
  \centering
  \includegraphics[width=1.0\linewidth]{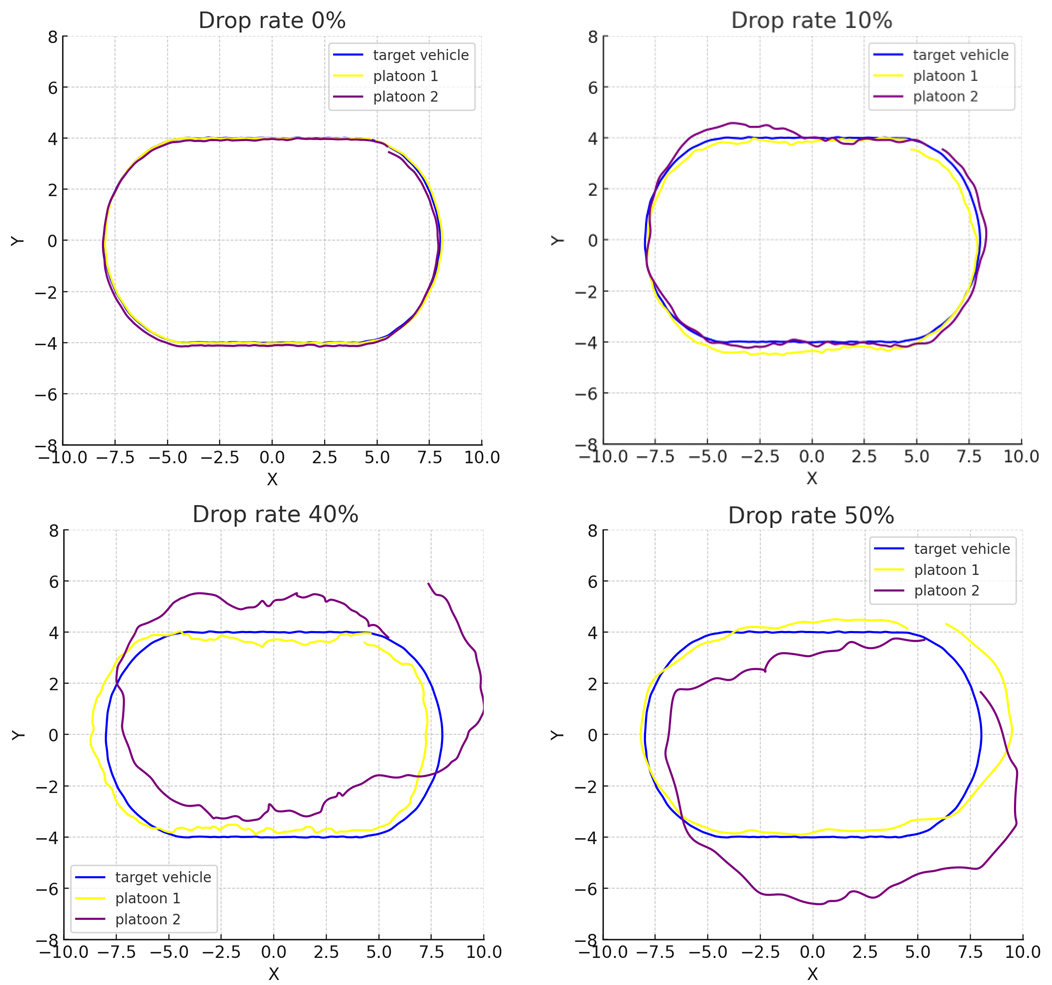}
  \caption{Plan-View Trajectories Under V2V Packet Drop Rates.}
  \label{fig:result}
\end{figure}

\subsection{Three-Vehicle Convoy Experiment}
To demonstrate multi-vehicle capability, the same track was traversed by a three-car convoy (Fig.~\ref{fig:result}).  
Each follower received Basic Safety Messages (BSMs) from \emph{all} upstream vehicles and ran the adaptive time-gap controller implemented in the ConvoyNext stack~\cite{maghsoumi2025convoynext}.  
Controller gains and communication cadence were identical across cars.

\section{Conclusions}

In this paper, we presented a detailed overview of the F1SIXTH platform, outlining its mechanical and electronic architecture, documenting specific challenges, and demonstrating how targeted upgrades can substantially improve reliability. We also introduced a concise step-by-step setup guide, along with essential firmware configurations for autopilot systems, ensuring that researchers can quickly adapt F1SIXTH to their experimental needs. Finally, we made our custom holder design publicly available, allowing others to integrate sensors and autopilot hardware with minimal modifications. We hope this work facilitates more accessible, reproducible, and scalable testing of autonomous driving algorithms in real-world-like settings.

\bibliographystyle{IEEEbib}
\bibliography{refs}

\end{document}